\pdfoutput=1
\documentclass[11pt]{article}

\usepackage{acl}

\usepackage{times}
\usepackage{latexsym}

\usepackage[T1]{fontenc}
\usepackage[utf8]{inputenc}

\usepackage{microtype}

\usepackage{inconsolata}

\usepackage{xspace}
\usepackage{amsmath}
\usepackage[compact]{titlesec}

\usepackage{graphicx}

\usepackage{tabularx}
\newcommand{\doublespace}[1]{\xspace\xspace}

\newcommand{\ns}[1]{\textcolor{black}{#1}}

\newcommand{\rev}[1]{\textcolor{black}{#1}}

\newcommand{\gptfrozen}[0]{GPT-2{\tiny{frozen}}}
\newcommand{\gptlmft}[0]{GPT-2{\tiny{HLC}}\xspace}% GPT2 finetuned on HuLM-Corpus

\newcommand{\hart}[0]{HaRT\xspace}
\newcommand{\hulm}{\textsc{HuLM}\xspace}

\newcommand{\name}[0]{GRIT\xspace}
\newcommand{\nameage}[0]{GRIT{\tiny{age}}\xspace}
\newcommand{\nameope}[0]{GRIT{\tiny{ope}}\xspace}

\newcommand{\bertds}[0]{BERT{\tiny{DS}}\xspace}
\newcommand{\bertagemlm}[0]{BERT{\tiny{age-MLM}}\xspace}

\title{Comparing Pre-trained Human Language Models:\\Is it Better with Human Context as Groups, Individual Traits, or Both?}

\author{Nikita Soni$^1$, Niranjan Balasubramanian$^1$, H. Andrew Schwartz$^1$, \and Dirk Hovy$^2$ \\
$^1$Stony Brook University,  $^2$Bocconi University \\
\texttt{\{nisoni, niranjan, has\}@cs.stonybrook.edu}, dirk.hovy@unibocconi.it}

\newcounter{auxFootnote}

\begin{document}
\maketitle
\begin{abstract}
\ns{Pre-trained language models consider the context of neighboring words and documents but lack any author context of the human generating the text. However, language depends on the author's states, traits, social, situational, and environmental attributes, collectively referred to as human context \citep{soni-etal-2024-large}.}
\rev{Human-centered natural language processing requires incorporating human context into language models}. Currently, two methods exist: \rev{pre-training with 1)} group-wise attributes (e.g., \textit{over-45-year-olds}) or \rev{2)} individual traits. Group attributes are \rev{simple but} coarse --- not all 45-year-olds write the same way --- while individual traits allow for more personalized representation\rev{s}, but require more complex modeling and data. \rev{I}t is unclear which approach benefits what tasks.
We compare pre-training models with human context via 1) group attributes, 2) individual users, and 3) a combined approach on five user- and document-level tasks.
%\rev{Using g}roup \textit{and} individual features together significantly improves \wassarev{the two} \textit{user}-level regression tasks like age estimation. \rev{Using} individual context significantly improves \wassarev{the three} \textit{document}-level classification tasks like stance detection. \rev{Combined individual and g}roup context reduce\rev{s} performance for document-level tasks.
Our results show that there is no best approach, but that human-centered language modeling holds avenues for different methods.
\end{abstract}

\section{Introduction}
Language \ns{is a fundamental form of human expression that} varies between people. \ns{Pre-trained Language Models (PLMs) account for the textual context of neighboring words and documents but lack the human context of the author ``generating" the language. However, language is highly dependent on the human context \cite{soni-etal-2024-large}, i.e., an author's changing states \cite{fleeson2001toward, mehl2003sounds}, traits, social, situational, and environmental attributes. For example, a person's language differs when hiking (situation/environment) versus when feeling dejected (state) over a breakup (situation). It is essential to model the additional \textit{human context} to better understand human language with PLMs \cite{soni-etal-2024-large}.} 
Two strands of human-centered Natural Language Processing (NLP) emerged to model the people behind the language. The first focuses on the \textit{group context}, building on the sociolinguistic notion of specific socio-demographic attributes influencing the language of a particular group. These socio-demographic attributes include age, gender \citep{volkova_exploring_2013, hovy_demographic_2015}, location \citep{kulkarni_freshman_2016, garimella_demographic-aware_2017}, personality \citep{schwartz_personality_2013, lynn_human_2017}, and more. 
The second approach focuses on building personalized language models that target \textit{individualistic contexts} \citep{king-cook-2020-evaluating, delasalles_learning_2019}, and latent attributes inferred from an individual's historical language~\citep{matero-etal-2021-melt-message, soni-etal-2022-human} to better model the user. 

\begin{figure}[!t]
    \centering
    \includegraphics[width=0.48\textwidth]{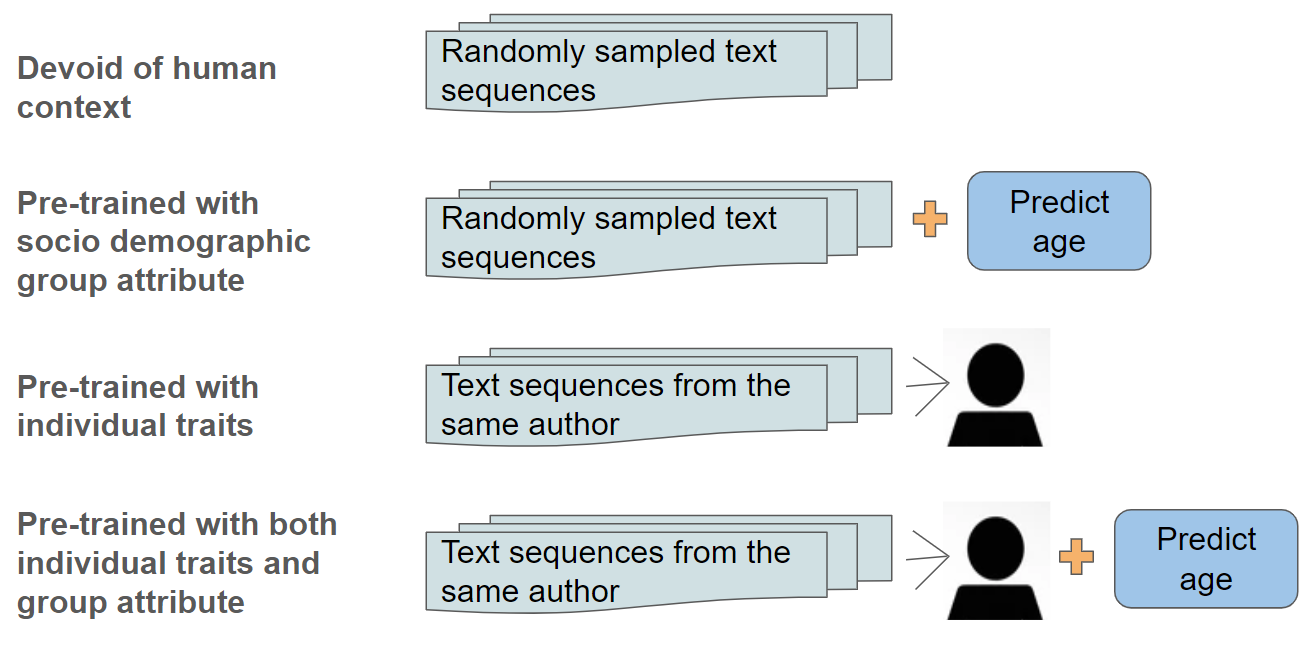}
    \caption{Pre-training a language model with no human context, with socio-demographic group attribute, with individual traits, and with both group and individual traits. }
    \label{fig:pt_humancontexts}
\end{figure}

While these two strands have advanced human-centered NLP, we still do not understand their relative strengths, complementarity, and impact on different tasks \citep{soni-etal-2024-large}. People are not defined by their group memberships alone \cite{orlikowski-etal-2023-ecological}, and individual traits might not be generalizable \rev{enough} across groups. \ns{Further, cross-cultural psychology research \citep{hofstede1984hofstede} notes the importance of both individualism and collectivism and \citeauthor{soni-etal-2024-large} suggest a flexible interplay of these concepts when including human context in PLMs.}  \rev{We might expect} models informed by both group and individual features to perform better\rev{, but no data exists on this}. In this work, we empirical\rev{ly} evaluate th\rev{ese} hypotheses and compare the effects of \rev{including} different types of human context in PLMs \ns{(i.e., groups context (collectivism), individualistic aspects (individualism), and a combination of both)}  o\rev{n} specific tasks.
We answer the following broader research questions:\\
\textit{\textbf{(RQ1)}: 
How can we incorporate group and individual human context into pre-training?}\\
\textit{\textbf{(RQ2)}: How does pre-training with different human contexts \rev{affect} downstream performance for different tasks?} 

Recent works trained PLMs with socio-demographic group context \citep{hung-etal-2023-demographic}, and individual context \citep{soni-etal-2022-human}. However, no empirical studies compare the impacts of different types of human contexts included in language modeling. We compare the downstream tasks' performance of models from these works with that of a new PLM trained with group and individual contexts. We test all systems on five \ns{downstream tasks from these works to ensure a variety of tasks and prediction properties at three levels:
1) user level, with historical language from authors (age estimation and personality assessment), 
2) document-level, with historical language from some authors (stance detection), 
3) document-level, without historical language from authors (topic detection and age category classification)}. 

Note that because we focus on empirically comparing 
pre-training with different types of human context, we cannot compare to the larger language models like GPT4, which are not yet pre-trained/trainable with human context \ns{but are envisioned to become large human language models in the future} \citep{soni-etal-2024-large}.
% \citep{soni2024large}.
Recent studies have explored methodologies like user-adapters \citep{zhong_useradapter_2021} and user-centric prompting \citep{li2023automatic} to include human context into the larger language models. In contrast, we focus specifically on comparing the impact of \textit{pre-training} LMs with different human contexts, \ns{as \citet{soni-etal-2024-large} argue that pre-training will allow for modeling a richer human context by explicitly handling the multi-level structure of documents connected to people, as opposed to specific and limited benefits of user-centric prompting and fine-tuning \citep{salemi2023lamp, choi-etal-2023-llms}.}

PLMs pre-trained on individual \textit{and} group features enhance performance on two user-level regression tasks that use multiple user documents\rev{:} age estimation and personality assessment. 
In contrast, PLMs pre-trained on \textit{individual} human context alone improve performance on document-level classification tasks like stance \rev{and topic} detection. 
Our findings suggest user-level tasks focus\rev{ing} on individual people are best modeled as a mix of both group attributes and unique characteristics. However, document-level tasks that are more personal\rev{, like stance detection,} are best addressed by modeling the individual context alone. Adding group attributes \rev{degrades} performance.

By their very nature, models of this kind touch upon sensitive user information. For this reason, we adopt a responsible release strategy, making only the code for the comparisons publicly available, along with the exact splits of the TrustPilot and Stance datasets used\footnote{\href{https://github.com/soni-n/HumanContextLanguageModels_Comparison}{https://github.com/soni-n/HumanContextLanguageModels\_Comparison}}. We build on top of the publicly available code from \citet{soni-etal-2022-human} and \citet{hung-etal-2023-demographic}. We acquired the models and data from the authors of the respective works in a secure manner. For more information about the models and data, see Sections \ref{model} and \ref{expmt_method}. For a discussion of the ethical implications of the models and data, see the Ethical Considerations section.

\paragraph{Contributions.} (1) We empirically compare three pre-training strategies for language models with human context: individual traits, group socio-demographic features, and a combination of both.

\noindent(2) We evaluate each pre-training strategy on five downstream tasks: two multi-document user-level regression (personality-openness evaluation and age estimation) and three single document-level classification tasks (stance detection, topic detection, and age category \ns{classification).}

\noindent(3) We find that the two user-level regression tasks perform better with PLMs pre-trained with individual and group human contexts. Conversely, the three single document-level tasks perform better with PLMs pre-trained with individual context alone. Further, pre-training with group and individual contexts reduces performance for document-level tasks.

\section{Related Work}
\paragraph{Socio-demographic and latent human factors.}Much work in human-centered NLP has focused on identifying and evaluating inclusion of human context in our models.
Initial studies show benefits of grouping people by socio-demographic factors like age or gender \citep{volkova_exploring_2013, hovy_demographic_2015} and geographical region \citep{bamman_distributed_2014, garimella_demographic-aware_2017} to capture the variation in language usage and meaning among different groups, and improving text classification tasks like sentiment analysis.
Additionally, adapting to socio-demographic user factors \citep{lynn_human_2017}, social networks \citep{huang-etal-2014-enriching, radfar_characterizing_2020}, and social media attributes \citep{bamman_contextualized_2015} have been effective to enhance the performance in tasks like sarcasm detection, and toxic language detection. Some studies go beyond modeling explicit groups, to learn individual representations latently \citet{jaech-ostendorf-2018-personalized, delasalles_learning_2019} or via historical language \citet{matero-etal-2021-melt-message}.  

\paragraph{Pre-traing with human context.}With respect to pre-trained LMs, recent studies have used adapter-based methodology \citep{li2021personalized, zhong_useradapter_2021} to include individual human contexts for downstream tasks. More recently, large language models have used user-centric prompting \citep{li2023automatic} to include human context and evaluate on personalized and social tasks, resulting in mediocre performance \citep{salemi2023lamp, choi-etal-2023-llms}. However, few studies have explored including human context within the pre-training regime of LMs. \citet{hung-etal-2023-demographic} generalize the task-specific EMPATH-BERT \citep{guda-etal-2021-empathbert} to create a PLM injected with demographic group information using a dynamic multi-task learning setup. Additionally, \citet{soni-etal-2022-human} pre-train a LM with individual human context derived from user's historical language. Our study aims at comparing the impacts of pre-training LMs with individual, or group, or combined individual plus group human context.

\section{Integrating Human Context in PLMs}
For our comparison, we use three systems representing the three paradigms of pre-training with human context (Figure \ref{fig:pt_humancontexts}).
We want to tease apart the contributions of different types of human context: 1) grouping people, \noindent2) modeling individual users, and 3) modeling both group and individual human contexts. As noted earlier, we focus on recent approaches for pre-training language models with additional human context.

\paragraph{Pre-training with group context.} 
We build on a model from \citet{hung-etal-2023-demographic} that explores demographic adaptation in transformer-based PLMs. It is a bidirectional auto-encoder-based PLM injecting demographic knowledge in a multi-task learning setup where they train for masked language modeling (MLM) and classify the gender or age of an author. They use the Trustpilot dataset \footnote{https://www.trustpilot.com/} of multilingual reviews with demographic labels \cite{hovy_demographic_2015}, and evaluate on multiple text classification tasks, including demographic attribute classification, sentiment analysis, and topic detection. For our comparison study, we use the US-English subset of the Trustpilot data for two tasks: topic detection (TD) across two age categories, and age attribute classification (AC) (more details in section \ref{expmt_method}). Additionally, we use a monolingual BERT pre-trained with age specialization on the Blogs authorship corpus \cite{schler2006effects}. This choice \ns{allows} us \ns{to} eliminate \ns{a} domain influence (i.e., Trustpilot reviews), given that the other PLMs \ns{under} compar\ns{ison} lack this specialization. 

\paragraph{Pre-training with individual human context.} 
\citet{soni-etal-2022-human} introduced human language modeling (HuLM) in PLMs, which is regular language modeling given an additional context of the individual generating the language. This additional context is a dynamic vector derived from the authors' historical texts motivated by the idea of capturing the changing human states expressed in language. It also adds coherence to texts generated by the same author. They introduce Human-aware Recurrent Transformer (HaRT), an autoregressive PLM to evaluate the effect of individual human context on language modeling and multiple user-level and document-level downstream tasks. We build on HaRT and use two user-level tasks, age estimation and personality (openness) assessment, and on a document-level task, stance detection, for our comparisons study.

\paragraph{Pre-training with both group and individual human context.} 
We train a PLM to integrate both individual and group human context \ns{by} introduc\ns{ing} a multi-task learning setup into HaRT \ns{that incorporates} group features. This \ns{approach} facilitates training a PLM with both group and individual context. We evaluate the model on two multi-document user-level regression tasks: age estimation and personality assessment, and three single document-level classification tasks: stance detection, topic detection, and age group classification. \ns{Importantly, the only difference in this multi-task learning setup compared to \hart is the inclusion of a demographic attribute prediction during pre-training, similar to how \citet{hung-etal-2023-demographic} adapted traditional PLMs for group context.}
% \andy{Importantly, the only difference between HaRT and this new model (GRiT) is whether it was pretrained with the multi-task setup to predict individual-level factors.}

\section{Models}
\label{model}

\subsection{Pre-training with individual human context}
\label{hart_and_hulm}

\paragraph{\hart.}\citet{soni-etal-2022-human} use a 12-layered autoregressive GPT-2 based architecture with a modified self-attention computation at layer 2. This modification to the query vector now includes the individual human context via a dynamic user-state vector.
\begin{equation*}
\it{Q^{IN}_{i}} = \it{W_q^T}[\it{H^{(IN-1)}_i};\it{U_{i-1}}] 
\end{equation*}
where $IN$ is the insert layer (layer 2), $Q_{i}$ is the query vector under computation, $H_i$ is the hidden states vector, and $U_{i-1}$ is the user-state vector derived from the previous block of language seen from the user. All the text from a user is processed in the same forward pass with recurrent processing of blocks of fixed-length (1024) tokens chunked after temporally ordering the social media posts by created time.
The user state is recurrently updated using the hidden states from layer 11 and computed as follows:
\begin{equation*}
U_{i} = tanh(W_{U} U_{i-1} + W_H H^{(E)})
\end{equation*}
where, $E$ is the extract layer (layer 11), $U_{i}$ is the updated user-state vector,  $U_{i-1}$ is the user-state vector from the previous block, and $H^{E}$ is the hidden states vector from layer 11. This formulation of updating the user-state vector extends the previous user-state vector information with the current language block's information.

\paragraph{\hulm Pre-training Task.}\hart is pre-trained for the human language modeling (\hulm) task defined as predicting the next token given the previous tokens while conditioning on previous user state $U_{1:t-1}$  \citep{soni-etal-2022-human} .
\begin{equation*}
Pr(\mathbf{W}_t|\mathbf{U}_{t-1}) = \prod_{i=1}^n Pr(w_{t,i} | w_{t,1:i-1}, \mathbf{U}_{1:t-1})
\end{equation*} 

This is translated into a pre-training objective to maximize:
\begin{equation*}
\prod_{a \in \textup{Users}}
\prod_{t = 1}^{|\mathcal{B}_a|} \prod_{i=1}^{|B_{t}^{(a)}|} Pr(w_{t,i} | w_{t,1:i-1}, B_{1:t-1}^{(a)})
\end{equation*}
where, $w_{t,i}$ is the $i^{th}$ token in the $t^{th}$ block ($B_{t}^{(a)}$) for user $a$. The tokens from the previous blocks are represented using \hart's recurrently updated user-state vector.
\citeauthor{soni-etal-2022-human} use cross-entropy loss for the \hulm objective.
\subsection{Pre-training with group human context}

\paragraph{\bertds and \bertagemlm.} \citet{hung-etal-2023-demographic} explore socio-demographic adapted BERT models to inject group human context into PLMs. We use two models: one specialized for age (demographic attribute) under the multi-task learning setup, and the other adapted to the age corpora through standard masked language modeling. We denote these as \bertds and \bertagemlm, respectively.

\paragraph{Multi-Task Learning.} \citet{hung-etal-2023-demographic} train for both domain adaptation using the masked language modeling ($L_{mlm}$) loss and for classifying demographic category using the binary cross-entropy loss ($L_{dem}$). Both losses must be combined to simultaneously learn multiple objectives.
To account for the \textit{homoscedastic uncertainty} \cite{kendall2018multi} of both losses, they adopt a dynamic multi-task learning (MTL) objective for training with group human context. Homoscedastic uncertainty is a task-dependent weighting to derive a multi-task loss function that can optimally learn the weights and balance the impact of multiple loss functions and their different scales. The tasks are dynamically weighted using the variance of the task-specific loss ($\sigma_{t}^2$) over training instances of the task $t \in \{mlm, dem\}$:
\begin{equation*}
\tilde{L_{t}} = \frac{1}{2\sigma_{t}^2} L_{t} + \log \sigma_{t}
\end{equation*}
\citeauthor{hung-etal-2023-demographic} minimize the sum of both the uncertainty adjusted losses: $\tilde{L}_{mlm} + \tilde{L}_{dem}$.

\subsection{Pre-training with both individual and group human context}
\label{grit}

\paragraph{\name.}
We train HaRT under a multi-task learning setup for both the individual context \textemdash{} through the HuLM pre-training task (see Section \ref{hart_and_hulm}) \textemdash{} and the group features \textemdash{} via a regression task to predict a (continuous) socio-demographic attribute of the author. We call the model as
\textbf{GR}oup and \textbf{I}ndividual HaR\textbf{T} (\name).
The model uses the user-state vectors (see Section \ref{hart_and_hulm}) to predict the socio-demographic attribute of the author:
\begin{equation*}
Pr(attribute|\overline{\mathbf{U}})
\end{equation*}

We chunk a user's language history into blocks and process them in a single forward pass. Each block of text from a user results in a user-state vector. We use the average of the user-state vectors from each non-padded block of texts from an author to compute their final user-state representation. This representation is layer-normed and linearly transformed before making a continuous-valued prediction for the specific attribute. 

We pre-train one model for the continuous attribute age (\nameage) and one for the continuous attribute personality type openness (\nameope). The models train on a regression loss for the attribute prediction regression tasks using mean squared error loss ($L_{mse}$), and a classification loss for the \hulm task using cross-entropy loss ($L_{ce}$). We must combine both losses to jointly learn the two objectives and account for the \textit{homoscedastic uncertainty} \cite{kendall2018multi} of the losses. Since we combine a regression and a classification loss, we train the model to learn to balance the loss for a continuous and discrete output as derived in \citet{kendall2018multi} and compute our joint objective as follows:
\begin{equation*}
\frac{1}{\sigma_{ce}^2} L_{ce} + \frac{1}{2\sigma_{mse}^2} L_{mse} + \log \sigma_{ce} + \log \sigma_{mse}    
\end{equation*}

\noindent
where, $\sigma_{ce}^2$ and $\sigma_{mse}^2$ are the variances of the task-specific losses over the training instances of the respective tasks.

To add numerical stability, we adjust the loss calculation to use log of the variance:
\begin{equation*}
\exp^{- \eta_{ce}} L_{ce} + \frac{1}{2} (\exp^{-\eta_{mse}} L_{mse} + \eta_{ce} + \eta_{mse})
\end{equation*}

\noindent
where $\eta_x = \log \sigma_{x}^2$ for $x \in \{mse, ce\}$. We let $\sigma_{ce}$ and $\sigma_{mse}$ be learnable parameters for the model.
In practice, we do not halve the $\eta_{ce}$ term in the above equation since we found it to perform better with our multi-task learning experiments.

\paragraph{Pre-training Data and Training.}
We use the same Facebook posts dataset \citep{park_automatic_2015} and training, validation, and test splits as those used by \citet{soni-etal-2022-human}. For both \nameage and \nameope, we use the demographic and personality scores, respectively, obtained from consenting Facebook users \citep{Kosinski5802}. This data is identical to that used by HaRT for the age estimation and personality assessment tasks. During training, we use a learning rate of 5e-5 in the multi-tasking training setup, employing the homoscedastic loss computation method described earlier. Following the experimental settings for \hart, each training instance is capped to 4 blocks of 1024 tokens each. We use a train batch size of 1 per device and an evaluation batch size of 20 per device, trained over 2 GPUs for 8 epochs. Further details can be found in Appendix \ref{PT_grit_appdx}.

\subsection{Fine-Tuning}

We utilize the results of fine-tuned \bertds and \bertagemlm from \citet{hung-etal-2023-demographic}, as well as fine-tuned \hart models from \citet{soni-etal-2022-human} where available. We fine-tune both GRIT models for all downstream tasks, and \hart for 2 document-level tasks. Additionally, we use the Optuna framework \cite{akiba2019optuna} for hyperparameter search, closely following the experimental settings in \citet{soni-etal-2022-human}. Details can be found in Appendix \ref{expmt_settings}.

\subsection{Transfer Learning}
\label{transfer_learn}
We experiment with fine-tuning \nameage in a multi-task learning setup for both the \hulm task and predicting personality (openness). Similarly, we fine-tune \nameope to predict age while also training for the \hulm task. We observe that this form of transfer learning yields the best performance for the user-level regression tasks (refer to Section \ref{results_comparison}).

\section{Experiments}
\label{expmt_method}
Our study's goal is to compare the downstream performance of models pre-trained with human contexts in three forms: socio-demographic group factors, individual traits, and combined.
To this end, we evaluate performances of the models defined in Section \ref{model} on two multi-document user-level regression tasks: predicting age and a personality score (openness), and on three single document-level classification tasks: stance detection, topic detection, and age classification. We also compare against \gptlmft from \citet{soni-etal-2022-human} as a PLM adapted to the social media domain but devoid of human context. \ns{All experiments were run using Optuna trials \citep{akiba2019optuna} to search for the best hyperparameters and reduce the effects of randomness. More details are provided in Appendix \ref{expmt_settings}}

\subsection{User Level Regression Tasks}
We consider two user-level social scientific tasks: age estimation, and personality (openness) assessment, which require predicting continuous outcomes (real-valued age, or openness score) for a user given multiple documents written by them. We use the same data splits as used by \citet{soni-etal-2022-human} for our comparison study.

Since \nameage is pre-trained using age estimation as one of the tasks, we use directly evaluate it on the held-out test set. This allows for direct comparison with \hart fine-tuned for the age estimation task. Furthermore, we can potentially attribute performance differences to the training with combined group and individual context, as \nameage incorporates the group feature into \hart's architecture. Similarly, \nameope is evaluated on the held-out test set for personality assessment. Moreover, we evaluate \nameage and \nameope for the tasks of personality assessment and age estimation, respectively, using the transfer learning mechanism described in Section \ref{transfer_learn}.
We report and compare pearson $r$ for age estimation and disattenuated pearson $r$ for personality assessment.

\subsection{Document-Level Classification Tasks}
\label{hungbert}
We compare different models for stance detection vs.\ topic detection and age classification tasks.
These tasks classify a single input document (tweet message or a review) that a user writes into label categories. For stance detection, we also utilize the historical messages of a user where available, as in \citet{soni-etal-2022-human}. However, we do not have the user information or any user historical language available for the other two tasks, so we evaluate solely based on the single document input. 

All models process the input document(s) and feed the layer-normed last non-padded token representation to the classification layer to classify the document into label categories. Only \name and \hart incorporate user information and the historical language available for the stance detection task. However, \gptlmft, and both \bertds and \bertagemlm lack this hierarchical structure and can only use the input document without access to historical data for making predictions. We compare the results from \citet{soni-etal-2022-human} and \citet{hung-etal-2023-demographic} wherever applicable and fine-tune all the parameters of the respective pre-trained models and the classification heads for other task-model combinations using the standard cross-entropy loss.

\paragraph{Stance Detection} 
Given a single annotated tweet, this task predicts a user's stance as in favor of, against, or neutral towards one of the five targets: atheism, climate change as a real concern, feminism, Hillary Clinton, and legalization of abortion. We fine-tune \ns{\nameage and \nameope} 
% the models under comparison 
for each target separately\ns{, and use the results from \citet{soni-etal-2022-human} for \gptlmft and \hart}.
We report the average of weighted F1 scores\footnotemark\setcounter{auxFootnote}{\value{footnote}} with three labels across all five targets. We use the train/dev/test split provided by \citet{soni-etal-2022-human} over the SemEval 2016 dataset \citep{StanceSemEval2016}. \hart and \name models maintain the temporal accuracy by using only the messages posted earlier than the labeled messages from the extended dataset \citep{lynn_tweet_2019} as a user's historical language.
% We report and compare the weighted F1 scores\verb|\footnote{\label{consistent_metrics}To be consistent with metrics from previous work.} of fine-tuned \gptlmft, \hart \citep{soni-etal-2022-human}, and fine-tuned \nameage and \nameope.
\footnotetext{We use this metric to maintain consistency with previous works under comparison \citep{soni-etal-2022-human, hung-etal-2023-demographic}.}

\paragraph{Topic Detection}
We use the US subset of the TrustPilot reviews dataset \cite{hovy_demographic_2015} from two age groups: below 35 or above 45 \footnote{As suggested by \citet{hovy_demographic_2015}, this split of the age ranges results in roughly equally-sized data sets and is non-contiguous, avoiding fuzzy boundaries.}. Given a single review, the task is to predict the review topics from five categories: Flights, Online marketplace, Fitness \& Nutrition, Electronics, and Hotels. To maintain consistency, we adopt the same train, development, and test set splits as \citet{hung-etal-2023-demographic} to ensure a stratified demographically-conditioned label distribution. We fine-tune \gptlmft, \hart, \nameage, and \nameope using these data splits to predict the topic for a given review, and report macro-F1 scores\footnotemark[\value{auxFootnote}]. We also compare to results from \bertagemlm and \bertds \citep{hung-etal-2023-demographic}.

\paragraph{Demographic Attribute Classification} 
We use the same subset of the TrustPilot dataset as for topic detection and the same train, development, and test splits from \citet{hung-etal-2023-demographic}. Given a single review, this task predicts the age group binary label (<35 years old or >45 years old). Age categories are equally represented in each set. We fine-tune \gptlmft, \hart, \nameage and \nameope using the provided splits to predict if the review is written by someone below 35 years or above 45 years, and report macro-F1 scores\footnotemark[\value{auxFootnote}]. We also compare to results from \bertagemlm and \bertds \citep{hung-etal-2023-demographic}.
\begin{table}[t]
\begin{tabularx}{0.48\textwidth} {
| >{\raggedright\arraybackslash}X
| >{\raggedright\arraybackslash}X 
| >{\centering\arraybackslash}X 
| >{\centering\arraybackslash}X |}
\hline 
\bf Model & \bf Human Context & \it Age ($r$) & \it OPE ($r_{dis}$) \\
\hline
\gptlmft & None & 0.839 & 0.521 \\
\hart & Individual & 0.868 & 0.619 \\
\nameage & Ind + Grp  & \bf 0.890 & \bf 0.658 \\
\nameope & Ind + Grp & 0.884 & 0.643 \\
\hline
\end{tabularx}

\caption{\label{tab:user-level} 
Pearson $r$ for age, disattenuated Pearson $r$ for openness. Pre-training with individual plus group context show benefits in estimating age and assessing personality (openness). Bold = best in column. We find no statistical difference between \nameage and \nameope for the task of age estimation. All other results show statistical significance $p < 0.05$ using paired t-test.
}

% paired t-test on error of predictions:
% Grit_age_err = abs(y - ypred_age) #vector algebra
% Grit_ope_err = abs(y - ypred_opee) #vector algebra
% ^both vectors in the same order
% paired_ttest(grit_age_err, grit_ope_err)

\end{table}

\subsection{Human Language Modeling}
To compare the effects of individual and group factors on language modeling performance, we evaluate on the test set from the pre-trained data splits. We report and compare perplexity scores from the pre-trained GPT-2 (\gptfrozen), \gptlmft, \hart, \nameage and \nameope for the human language modeling task.

\begin{table*}
\centering
\begin{tabularx}{0.85\textwidth} {
| >{\raggedright\arraybackslash}X
| >{\raggedright\arraybackslash}X 
| >{\centering\arraybackslash}X 
| >{\centering\arraybackslash}X 
| >{\centering\arraybackslash}X 
| >{\centering\arraybackslash}X |}
\hline 
\bf Model & \bf Human Context & \it Stance (F1{\tiny{wtd}}) & \it TD (<35) (F1{\tiny{mac}}) & \it TD (>45) (F1{\tiny{mac}})  & \it AC \doublespace{}\doublespace{}(F1{\tiny{mac}}) \\  

\hline
\gptlmft & None & 68.60 & 69.77 & 65.43 & 63.93 \\

\bertagemlm & Group & - & 68.40 & 64.60 & 61.90  \\

\bertds & Group & - & 69.30 & 65.00 & 64.10  \\

% \hart & Individual & \space$\textbf{\doublespace{}71.10}^{*}$ & $\textbf{\doublespace{}69.84}^{*}$ & $\textbf{\doublespace{}65.65}^{*}$ & $\textbf{\doublespace{}64.33}^{*}$ \\

\hart & Individual & \space$\textbf{71.10}$ & $\textbf{69.84}$ & $\textbf{65.65}$ & $\textbf{64.33}$ \\

\nameage & Ind + Grp & 70.82 & 69.21 & 64.52 & 62.56 \\
\nameope & Ind + Grp & 70.07 & 66.53 & 64.84 & 61.18 \\
\hline
\end{tabularx}
\caption{\label{tab:doc-level} 
Weighted F1 for stance detection, macro-F1 for topic detection (TD), and age classification (AC) on TrustPilot reviews. Pre-training with individual context appear to benefit all tasks. \textbf{Bold} $=$ best in column; McNemar's test comparing classifiers does not show statistical significance between the best performing model (\hart) and the best baseline with no individual context (\gptlmft). }
\end{table*}

\section{Results and Discussion}
We report results for all the tasks here, discussing their respective impacts from pre-training LMs with individual human context, group context, and both individual and group context.

\subsection{Comparisons Study}
\label{results_comparison}
\paragraph{User-Level Regression Tasks.} Table \ref{tab:user-level} shows the results of the two user-level regression tasks.
% \andy{Importantly, HaRT and GRiT, only differ in that GRiT had the explicit factors as part of the pertaining task, so differences in performance can be attributed to that change.}
We find that \name models outperform others for both age estimation and personality assessment tasks.
Additionally, upon comparing the transfer learning (Section \ref{transfer_learn}) outcomes of \nameage for openness and  \nameope for age to those of the \hart and \gptlmft models, we consistently observe superior performance with the \name models, further substantiating their efficacy.

Note that while \gptlmft is a PLM that is adapted to the social-media domain, it \ns{lacks} human context. \hart \ns{incorporates} individual human context \ns{in pre-training}, and \name \ns{extends this by integrating both} group and individual human contexts \ns{in pre-training} (Figure \ref{fig:pt_humancontexts}). As shown in Table \ref{tab:user-level}, there are gains \ns{observed} from \gptlmft (no human context) to \hart (individual human context), and further to \name (individual + group human context). This \ns{suggests} that pre-training PLMs with individual and group human context \ns{can} benefit multi-document user-level regression tasks, \ns{such as those} we considered. \ns{Importantly, the only difference between \hart and \name models lies in the integration of the demographic attribute prediction (group context). Both models are pre-trained and evaluated on precisely the same data, allowing performance differences to be attributed to the additional group context combined with individualistic human context.}

\paragraph{Document-Level Classification Tasks.}
Table \ref{tab:doc-level} shows the results for the 3 document-level classification tasks: stance detection, topic detection (TD) for 2 age groups (<35 and >45), and demographic attribute (age) group classification (AC). We see that task fine-tuned \hart (individual human context) models perform better on all tasks. 

\hart models inherently include an additional context of the individual user and do not treat all inputs as if written by the same user. The considered stance detection task primarily relates to personal opinions and preferences, rather than group-level ones, making \hart well-suited for incorporating such personalization due to its pre-training with individual human context. While a group context may also influence a person's stance to some extent, empirical observations show that the combination of individual and group contexts negatively impacts performance. Additionally, models pre-trained with group context (\bertds) perform well in group-based tasks such as topic detection and age classification. However, models pre-trained on both individual and group human context (\name) do not appear to enhance results in group-based, and personal stance detection tasks resulting in slightly worse performance.

Further, it is important to note that the individual human context (\hart) derived \ns{for some of the users using their}  historical tweets\ns{, where available,} in the stance detection dataset provides a richer human context \ns{as we see greater gains in the performance of \hart over \gptlmft. Conversely, when historical language is not available for certain datasets (topic detection and attribute classification), \hart does not perform worse than \gptlmft and may even achieve marginal gains due to the inherent human context in the model. However, we leave the evaluation of the impact of historical language on human context for future work.}
% the performance is not greatly hurt even if historical language is not available for certain tasks (TD and AC).

\begin{table}[t]
\begin{center}
\begin{tabularx} {0.4\textwidth} {
| >{\raggedright\arraybackslash}X 
| >{\raggedright\arraybackslash}X 
| >{\centering\arraybackslash}X |
}
\hline \bf Model & \bf Human Context &  Test  ($ppl$) \\
\hline
\gptfrozen & None & 114.82 \\
\gptlmft & None &  \text{\doublespace{}36.39}  \\
\hart & Individual & \textbf{\doublespace{}28.24} \\
% \nameage (4blocks)  &  31.88 \\
% \nameope (4blocks)  &  30.50 \\
\nameage & Ind + Grp &  \text{\doublespace{}31.77} \\
\nameope & Ind + Grp &  \text{\doublespace{}30.32} \\
\hline
\end{tabularx}
\end{center}
\caption{\label{tab:ppl}
Comparing perplexity on language modeling for models trained with individual and group contexts.}
\end{table}

\paragraph{Perplexity.}
We also compare the language modeling capability of the various models. Table \ref{tab:ppl} reports perplexity on the held-out test set. The frozen GPT-2 performs poorly compared to the social media domain adapted \gptlmft, while \hart model with individual human context perform the best. In contrast, \name models with both individual and group human context exhibit a slightly lower perplexity than \hart. An individual's language is inherently personal, yet it can also be influenced by their group context to some extent, thereby affecting the perplexity results in language modeling tasks. However, \name models pre-trained with both individual and group context yield slightly worse perplexity measures. Additionally, we observe similar trends in perplexity gains from \gptlmft (no human context) to \hart (individual context) or \name (individual plus group context) as also demonstrated in \citet{soni-etal-2022-human}.

\begin{table}
    \begin{center}
    
    \begin{tabularx}{0.49\textwidth} {
    | >{\hsize=0.8\hsize\raggedright\arraybackslash}X
    | >{\hsize=0.8\hsize\centering\arraybackslash}X 
    | >{\hsize=0.8\hsize\centering\arraybackslash}X 
    | >{\hsize=1.3\hsize\centering\arraybackslash}X 
    | >{\hsize=1.3\hsize\centering\arraybackslash}X |
    }
    \hline 
    \bf Age bucket & \#Users & \it{\hart} (Ind) & \it{\nameage} (Ind+Grp) & \it{\nameope} (Ind+Grp) \\ 
    \hline
    
    <18 & 1113 & 0.223 & \bf 0.394 & 0.393  \\
    18-21 & 1387 & 0.230 & \bf 0.278 & 0.276  \\
    21-30 & 1557 & 0.512 & \bf 0.531 & 0.519  \\
    30-45 & \text{\doublespace{}695} & 0.485 & \bf 0.530 & 0.520  \\
    45+ & \text{\doublespace{}248} & 0.106 & \bf 0.205 & 0.180  \\
    
    \hline
    \end{tabularx}
    \end{center}
    \caption{ 
        Pearson $r$ for age over five age buckets using different types of human contexts for error analysis. Bold indicates best in row. We find no statistical difference between \nameage and \nameope for buckets 21-30 and 30-45. All other results show statistical significance $p < 0.05$ using paired t-test.
        }
        \label{tab:error-analysis}
\end{table}

\begin{table}
    \begin{center}
    \begin{tabularx} {0.48\textwidth} {
| >{\hsize=1.3\hsize\raggedright\arraybackslash}X
| >{\hsize=0.7\hsize\centering\arraybackslash}X
| >{\centering\arraybackslash}X
| >{\centering\arraybackslash}X |
}
    \hline \bf Task\textbackslash Model & \it{\hart} (Ind) & \it{\nameage} (Ind+Grp) & \it{\nameope} (Ind+Grp) \\ 
    \hline
    
    Age ($r$) & 0.215 & \textbf{0.181} & 0.185  \\
    OPE ($r_{dis}$)  & 0.075 & 0.090 & \textbf{0.072}  \\
    
    \hline
    \end{tabularx}
    \end{center}
    \caption{\label{tab:error-disp} 
        Mean error disparity for age estimation and openness personality assessment over five age buckets. 
        Bold indicates best in column (lower is better). 
        }
\end{table}

\subsection{Error Analysis and Disparity}
We conduct an error analysis based on a socio-demographic group attribute (age groups), specifically focusing on age and openness prediction tasks. We measure the performance of GRIT and HaRT in terms of error disparity \citep{shah2020predictive} \textemdash{} a systematic difference in error based on demographics as exemplified by the ``Wall Street Journal Effect'' \cite{hovy_tagging_2015}. We analyze both the prediction outcomes and error disparity in age and openness prediction for both models: \hart, which considers individual context, and \name, which incorporates both individual and group context.

First, we split the task test dataset into different buckets based on the age groups (specifically, <18, 18-21, 21-30, 30-45, and >45 years old) of the users in the test set, and then we compare the performance of our models across these buckets. Results from Table \ref{tab:error-analysis} indicate that pre-training with individual and group contexts together performs better for estimating age across all the age groups, which implies it makes fewer errors as a function of the socio-demographic attribute age. We see similar trends for assessing openness personality (see Appendix Tables \ref{tab:error-analy-ope-r} and \ref{tab:error-analy-ope-mse}), suggesting that the group attribute prediction may act as a regularizer for models pre-trained with both individual and group contexts, thus aiding the models to make fewer errors across all age buckets.

To further confirm, we compute the mean error disparity ($MED$) as the sum of the differences in the performance metric (Pearson correlation for age, and disattenuated Pearson correlation for openness) across each pair of age buckets, which is then averaged by the number of pairs \citep{shah2020predictive}. A lower averaged sum of differences implies fewer errors as a function of the age groups. Lower $MED$ scores for models pre-trained with individual and group context in Table \ref{tab:error-disp} support our previous error analysis.

\section{Conclusion}
\label{sec: conclusion}
NLP benefits from modeling latent human context, such as socio-demographic group features or individual traits. 
A recent development has been to incorporate this additional human context into the pre-training regimen of LMs. However, humans exhibit varying degrees of group and individual characteristics. Understanding the impacts of pre-training with different types of human context will advance the integration of human context into our base LLMs \citep{soni-etal-2024-large}.
To assess the impacts, we compare three types of PLMs pre-trained with socio-demographic group attributes, individual human contexts, and combined group and individual traits, across five user- and document-level tasks. Our findings indicate that pre-training with both individual \textit{and} group human context improves the two user-level regression tasks: age and personality prediction. Pre-training with individual human context enhances the performance of the three single-document classification tasks, including stance and topic detection. Interestingly, inclusion of both individual and group attributes results in reduced performance on the text classification tasks. Meanwhile, pre-training solely on group context aids in group-based document classification tasks, albeit suboptimally. These results represent a promising step towards modeling human context and offer valuable insights for the NLP community to investigate additional strategies for improving models with task-dependent human context during pre-training.

\section*{Limitations}
\label{sec:limits}
The purpose of our study is to compare the impacts of modeling socio-demographic group attributes and modeling individual user traits, and we use relevant models to represent each of the approaches. There are likely to be other ways to model these approaches and the models we use are only one of the ways. Additionally, these models in themselves have limitations like the blocks mechanism to process all the text from author induces compute requirements resulting in a capping of the number of blocks used for training. While it is also unclear how many blocks are sufficient to capture the human context, and if it is helpful to use the earliest language or the most recently used language in the capped number of blocks.\\
Secondly, some of the datasets (TrustPilot) used do not have appropriate user identification or historical language to create an individual human context. \\
Lastly, as noted earlier, models and data that touch upon sensitive user information require an extremely responsible usage and limit researchers to make them publicly available.

\section*{Ethical Considerations}
\label{sec:ethics}
Models that incorporate socio-demographic information need to be considered with special scrutiny. On the one hand, they have the potential to produce fairer and more inclusive results, because they can account for human language variation. On the other hand, they risk revealing identifying or sensitive information, which can lead to profiling and stereotyping. These may present opportunities for unintended malicious exploitations. For example, models that improve demographic groups prediction or psychological assessments could be used for targeting content for individuals without their awareness or consent. Such models may also risk release of private information of the research participant if trained on private data unchecked for exposing identifying information. For this reason, we take a conservative release strategy. While we support open research and reproducibility, data and privacy protection take precedence. Thus, we will only be releasing the code for our comparison study and the data that does not contain sensitive information i.e., stance detection datasets and TrustPilot datasets for topic detection and attribute classification. This is also in accordance with the DUA we have received from the authors of the papers/models that we employ in our work. \\
Our comparison study aims to guide and further speed the growing body of human-centered AI research. The models under comparison aim to enable applicability in the interdisciplinary studies of the human condition leading to helpful tools for psychological health. However, at this point these models are not intended for use in practice and should be evaluated for failures. All user-level tasks presented here were reviewed and approved or exempted by an academic institutional review board (IRB). Our studies are limited to US-English due to comparability reasons. However, similar effects are likely to hold for other languages, and should be evaluated in future work.

\section*{Acknowledgments}
This research is supported in part by the Office of the Director of National Intelligence (ODNI), Intelligence Advanced Research Projects Activity (IARPA), via the HIATUS Program contract \#2022-22072200005, and a grant from the CDC/NIOSH (U01 OH012476). The views and conclusions contained herein are those of the authors and should not be interpreted as necessarily representing the official policies, either expressed or implied, of ODNI, IARPA, any other government organization, or the U.S. Government. The U.S. Government is authorized to reproduce and distribute reprints for governmental purposes notwithstanding any copyright annotation therein.

\bibliography{anthology,custom}

\begin{thebibliography}{36}
\expandafter\ifx\csname natexlab\endcsname\relax\def\natexlab#1{#1}\fi

\bibitem[{Akiba et~al.(2019)Akiba, Sano, Yanase, Ohta, and Koyama}]{akiba2019optuna}
Takuya Akiba, Shotaro Sano, Toshihiko Yanase, Takeru Ohta, and Masanori Koyama. 2019.
\newblock Optuna: A next-generation hyperparameter optimization framework.
\newblock In \emph{Proceedings of the 25th ACM SIGKDD international conference on knowledge discovery \& data mining}, pages 2623--2631.

\bibitem[{Bamman et~al.(2014)Bamman, Dyer, and Smith}]{bamman_distributed_2014}
David Bamman, Chris Dyer, and Noah~A. Smith. 2014.
\newblock \href {https://doi.org/10.3115/v1/P14-2134} {Distributed {Representations} of {Geographically} {Situated} {Language}}.
\newblock In \emph{Proceedings of the 52nd {Annual} {Meeting} of the {Association} for {Computational} {Linguistics} ({Volume} 2: {Short} {Papers})}, pages 828--834, Baltimore, Maryland. Association for Computational Linguistics.

\bibitem[{Bamman and Smith(2015)}]{bamman_contextualized_2015}
David Bamman and Noah Smith. 2015.
\newblock \href {https://ojs.aaai.org/index.php/ICWSM/article/view/14655} {Contextualized {Sarcasm} {Detection} on {Twitter}}.
\newblock \emph{Proceedings of the International AAAI Conference on Web and Social Media}, 9(1):574--577.
\newblock Number: 1.

\bibitem[{Choi et~al.(2023)Choi, Pei, Kumar, Shu, and Jurgens}]{choi-etal-2023-llms}
Minje Choi, Jiaxin Pei, Sagar Kumar, Chang Shu, and David Jurgens. 2023.
\newblock \href {https://doi.org/10.18653/v1/2023.emnlp-main.699} {Do {LLM}s understand social knowledge? evaluating the sociability of large language models with {S}oc{KET} benchmark}.
\newblock In \emph{Proceedings of the 2023 Conference on Empirical Methods in Natural Language Processing}, pages 11370--11403, Singapore. Association for Computational Linguistics.

\bibitem[{Delasalles et~al.(2019)Delasalles, Lamprier, and Denoyer}]{delasalles_learning_2019}
Edouard Delasalles, Sylvain Lamprier, and Ludovic Denoyer. 2019.
\newblock \href {https://doi.org/10.1109/ICDM.2019.00022} {Learning {Dynamic} {Author} {Representations} with {Temporal} {Language} {Models}}.
\newblock \emph{2019 IEEE International Conference on Data Mining (ICDM)}, pages 120--129.
\newblock ArXiv: 1909.04985.

\bibitem[{Fleeson(2001)}]{fleeson2001toward}
William Fleeson. 2001.
\newblock Toward a structure-and process-integrated view of personality: Traits as density distributions of states.
\newblock \emph{Journal of personality and social psychology}, 80(6):1011.

\bibitem[{Garimella et~al.(2017)Garimella, Banea, and Mihalcea}]{garimella_demographic-aware_2017}
Aparna Garimella, Carmen Banea, and Rada Mihalcea. 2017.
\newblock \href {https://doi.org/10.18653/v1/D17-1242} {Demographic-aware word associations}.
\newblock In \emph{Proceedings of the 2017 {Conference} on {Empirical} {Methods} in {Natural} {Language} {Processing}}, pages 2285--2295, Copenhagen, Denmark. Association for Computational Linguistics.

\bibitem[{Guda et~al.(2021)Guda, Garimella, and Chhaya}]{guda-etal-2021-empathbert}
Bhanu Prakash~Reddy Guda, Aparna Garimella, and Niyati Chhaya. 2021.
\newblock \href {https://doi.org/10.18653/v1/2021.eacl-main.268} {{E}mpath{BERT}: A {BERT}-based framework for demographic-aware empathy prediction}.
\newblock In \emph{Proceedings of the 16th Conference of the European Chapter of the Association for Computational Linguistics: Main Volume}, pages 3072--3079, Online. Association for Computational Linguistics.

\bibitem[{Hofstede and Bond(1984)}]{hofstede1984hofstede}
Geert Hofstede and Michael~H Bond. 1984.
\newblock Hofstede's culture dimensions: An independent validation using rokeach's value survey.
\newblock \emph{Journal of cross-cultural psychology}, 15(4):417--433.

\bibitem[{Hovy(2015)}]{hovy_demographic_2015}
Dirk Hovy. 2015.
\newblock \href {https://doi.org/10.3115/v1/P15-1073} {Demographic {Factors} {Improve} {Classification} {Performance}}.
\newblock In \emph{Proceedings of the 53rd {Annual} {Meeting} of the {Association} for {Computational} {Linguistics} and the 7th {International} {Joint} {Conference} on {Natural} {Language} {Processing} ({Volume} 1: {Long} {Papers})}, pages 752--762, Beijing, China. Association for Computational Linguistics.

\bibitem[{Hovy and Søgaard(2015)}]{hovy_tagging_2015}
Dirk Hovy and Anders Søgaard. 2015.
\newblock \href {https://doi.org/10.3115/v1/P15-2079} {Tagging {Performance} {Correlates} with {Author} {Age}}.
\newblock In \emph{Proceedings of the 53rd {Annual} {Meeting} of the {Association} for {Computational} {Linguistics} and the 7th {International} {Joint} {Conference} on {Natural} {Language} {Processing} ({Volume} 2: {Short} {Papers})}, pages 483--488, Beijing, China. Association for Computational Linguistics.

\bibitem[{Huang et~al.(2014)Huang, Yan, Kuo, and Lin}]{huang-etal-2014-enriching}
Yu-Yang Huang, Rui Yan, Tsung-Ting Kuo, and Shou-De Lin. 2014.
\newblock \href {https://doi.org/10.3115/v1/P14-2100} {Enriching cold start personalized language model using social network information}.
\newblock In \emph{Proceedings of the 52nd Annual Meeting of the Association for Computational Linguistics (Volume 2: Short Papers)}, pages 611--617, Baltimore, Maryland. Association for Computational Linguistics.

\bibitem[{Hung et~al.(2023)Hung, Lauscher, Hovy, Ponzetto, and Glava{\v{s}}}]{hung-etal-2023-demographic}
Chia-Chien Hung, Anne Lauscher, Dirk Hovy, Simone~Paolo Ponzetto, and Goran Glava{\v{s}}. 2023.
\newblock \href {https://aclanthology.org/2023.findings-eacl.116} {Can demographic factors improve text classification? revisiting demographic adaptation in the age of transformers}.
\newblock In \emph{Findings of the Association for Computational Linguistics: EACL 2023}, pages 1565--1580, Dubrovnik, Croatia. Association for Computational Linguistics.

\bibitem[{Jaech and Ostendorf(2018)}]{jaech-ostendorf-2018-personalized}
Aaron Jaech and Mari Ostendorf. 2018.
\newblock \href {https://doi.org/10.18653/v1/P18-2111} {Personalized language model for query auto-completion}.
\newblock In \emph{Proceedings of the 56th Annual Meeting of the Association for Computational Linguistics (Volume 2: Short Papers)}, pages 700--705, Melbourne, Australia. Association for Computational Linguistics.

\bibitem[{Kendall et~al.(2018)Kendall, Gal, and Cipolla}]{kendall2018multi}
Alex Kendall, Yarin Gal, and Roberto Cipolla. 2018.
\newblock Multi-task learning using uncertainty to weigh losses for scene geometry and semantics.
\newblock In \emph{Proceedings of the IEEE conference on computer vision and pattern recognition}, pages 7482--7491.

\bibitem[{King and Cook(2020)}]{king-cook-2020-evaluating}
Milton King and Paul Cook. 2020.
\newblock \href {https://aclanthology.org/2020.lrec-1.299} {Evaluating approaches to personalizing language models}.
\newblock In \emph{Proceedings of the Twelfth Language Resources and Evaluation Conference}, pages 2461--2469, Marseille, France. European Language Resources Association.

\bibitem[{Kosinski et~al.(2013)Kosinski, Stillwell, and Graepel}]{Kosinski5802}
Michal Kosinski, David Stillwell, and Thore Graepel. 2013.
\newblock \href {https://doi.org/10.1073/pnas.1218772110} {Private traits and attributes are predictable from digital records of human behavior}.
\newblock \emph{Proceedings of the National Academy of Sciences}, 110(15):5802--5805.

\bibitem[{Kulkarni et~al.(2016)Kulkarni, Perozzi, and Skiena}]{kulkarni_freshman_2016}
Vivek Kulkarni, Bryan Perozzi, and Steven Skiena. 2016.
\newblock \href {https://ojs.aaai.org/index.php/ICWSM/article/view/14798} {Freshman or {Fresher}? {Quantifying} the {Geographic} {Variation} of {Language} in {Online} {Social} {Media}}.
\newblock \emph{Proceedings of the International AAAI Conference on Web and Social Media}, 10(1):615--618.
\newblock Number: 1.

\bibitem[{Li et~al.(2023)Li, Zhang, Mei, Kong, and Bendersky}]{li2023automatic}
Cheng Li, Mingyang Zhang, Qiaozhu Mei, Weize Kong, and Michael Bendersky. 2023.
\newblock Automatic prompt rewriting for personalized text generation.
\newblock \emph{arXiv preprint arXiv:2310.00152}.

\bibitem[{Li et~al.(2021)Li, Zhang, and Chen}]{li2021personalized}
Lei Li, Yongfeng Zhang, and Li~Chen. 2021.
\newblock Personalized transformer for explainable recommendation.
\newblock In \emph{Proceedings of the 59th Annual Meeting of the Association for Computational Linguistics and the 11th International Joint Conference on Natural Language Processing (Volume 1: Long Papers)}, pages 4947--4957.

\bibitem[{Lynn et~al.(2019)Lynn, Giorgi, Balasubramanian, and Schwartz}]{lynn_tweet_2019}
Veronica Lynn, Salvatore Giorgi, Niranjan Balasubramanian, and H.~Andrew Schwartz. 2019.
\newblock \href {https://doi.org/10.18653/v1/W19-2103} {Tweet {Classification} without the {Tweet}: {An} {Empirical} {Examination} of {User} versus {Document} {Attributes}}.
\newblock In \emph{Proceedings of the {Third} {Workshop} on {Natural} {Language} {Processing} and {Computational} {Social} {Science}}, pages 18--28, Minneapolis, Minnesota. Association for Computational Linguistics.

\bibitem[{Lynn et~al.(2017)Lynn, Son, Kulkarni, Balasubramanian, and Schwartz}]{lynn_human_2017}
Veronica Lynn, Youngseo Son, Vivek Kulkarni, Niranjan Balasubramanian, and H.~Andrew Schwartz. 2017.
\newblock \href {https://doi.org/10.18653/v1/D17-1119} {Human {Centered} {NLP} with {User}-{Factor} {Adaptation}}.
\newblock In \emph{Proceedings of the 2017 {Conference} on {Empirical} {Methods} in {Natural} {Language} {Processing}}, pages 1146--1155, Copenhagen, Denmark. Association for Computational Linguistics.

\bibitem[{Matero et~al.(2021)Matero, Soni, Balasubramanian, and Schwartz}]{matero-etal-2021-melt-message}
Matthew Matero, Nikita Soni, Niranjan Balasubramanian, and H.~Andrew Schwartz. 2021.
\newblock \href {https://doi.org/10.18653/v1/2021.findings-emnlp.253} {{M}e{LT}: Message-level transformer with masked document representations as pre-training for stance detection}.
\newblock In \emph{Findings of the Association for Computational Linguistics: EMNLP 2021}, pages 2959--2966, Punta Cana, Dominican Republic. Association for Computational Linguistics.

\bibitem[{Mehl and Pennebaker(2003)}]{mehl2003sounds}
Matthias~R Mehl and James~W Pennebaker. 2003.
\newblock The sounds of social life: a psychometric analysis of students' daily social environments and natural conversations.
\newblock \emph{Journal of personality and social psychology}, 84(4):857.

\bibitem[{Mohammad et~al.(2016)Mohammad, Kiritchenko, Sobhani, Zhu, and Cherry}]{StanceSemEval2016}
Saif~M. Mohammad, Svetlana Kiritchenko, Parinaz Sobhani, Xiaodan Zhu, and Colin Cherry. 2016.
\newblock Semeval-2016 task 6: Detecting stance in tweets.
\newblock In \emph{Proceedings of the International Workshop on Semantic Evaluation}, SemEval '16, San Diego, California.

\bibitem[{Orlikowski et~al.(2023)Orlikowski, R{\"o}ttger, Cimiano, and Hovy}]{orlikowski-etal-2023-ecological}
Matthias Orlikowski, Paul R{\"o}ttger, Philipp Cimiano, and Dirk Hovy. 2023.
\newblock \href {https://doi.org/10.18653/v1/2023.acl-short.88} {The ecological fallacy in annotation: Modeling human label variation goes beyond sociodemographics}.
\newblock In \emph{Proceedings of the 61st Annual Meeting of the Association for Computational Linguistics (Volume 2: Short Papers)}, pages 1017--1029, Toronto, Canada. Association for Computational Linguistics.

\bibitem[{Park et~al.(2015)Park, Schwartz, Eichstaedt, Kern, Kosinski, Stillwell, Ungar, and Seligman}]{park_automatic_2015}
Gregory Park, H.~Andrew Schwartz, Johannes~C. Eichstaedt, Margaret~L. Kern, Michal Kosinski, David~J. Stillwell, Lyle~H. Ungar, and Martin E.~P. Seligman. 2015.
\newblock \href {https://doi.org/10.1037/pspp0000020} {Automatic personality assessment through social media language}.
\newblock \emph{Journal of Personality and Social Psychology}, 108(6):934--952.

\bibitem[{Radfar et~al.(2020)Radfar, Shivaram, and Culotta}]{radfar_characterizing_2020}
Bahar Radfar, Karthik Shivaram, and Aron Culotta. 2020.
\newblock \href {https://ojs.aaai.org/index.php/ICWSM/article/view/7366} {Characterizing {Variation} in {Toxic} {Language} by {Social} {Context}}.
\newblock \emph{Proceedings of the International AAAI Conference on Web and Social Media}, 14:959--963.

\bibitem[{Salemi et~al.(2023)Salemi, Mysore, Bendersky, and Zamani}]{salemi2023lamp}
Alireza Salemi, Sheshera Mysore, Michael Bendersky, and Hamed Zamani. 2023.
\newblock Lamp: When large language models meet personalization.
\newblock \emph{arXiv preprint arXiv:2304.11406}.

\bibitem[{Schler et~al.(2006)Schler, Koppel, Argamon, and Pennebaker}]{schler2006effects}
Jonathan Schler, Moshe Koppel, Shlomo Argamon, and James~W Pennebaker. 2006.
\newblock Effects of age and gender on blogging.
\newblock In \emph{AAAI spring symposium: Computational approaches to analyzing weblogs}, volume~6, pages 199--205.

\bibitem[{Schwartz et~al.(2013)Schwartz, Eichstaedt, Kern, Dziurzynski, Ramones, Agrawal, Shah, Kosinski, Stillwell, Seligman, and Ungar}]{schwartz_personality_2013}
H.~Andrew Schwartz, Johannes~C. Eichstaedt, Margaret~L. Kern, Lukasz Dziurzynski, Stephanie~M. Ramones, Megha Agrawal, Achal Shah, Michal Kosinski, David Stillwell, Martin E.~P. Seligman, and Lyle~H. Ungar. 2013.
\newblock \href {https://doi.org/10.1371/journal.pone.0073791} {Personality, {Gender}, and {Age} in the {Language} of {Social} {Media}: {The} {Open}-{Vocabulary} {Approach}}.
\newblock \emph{PLOS ONE}, 8(9):e73791.
\newblock Publisher: Public Library of Science.

\bibitem[{Shah et~al.(2020)Shah, Schwartz, and Hovy}]{shah2020predictive}
Deven~Santosh Shah, H~Andrew Schwartz, and Dirk Hovy. 2020.
\newblock Predictive biases in natural language processing models: A conceptual framework and overview.
\newblock In \emph{Proceedings of the 58th Annual Meeting of the Association for Computational Linguistics}, pages 5248--5264.

\bibitem[{Soni et~al.(2022)Soni, Matero, Balasubramanian, and Schwartz}]{soni-etal-2022-human}
Nikita Soni, Matthew Matero, Niranjan Balasubramanian, and H.~Andrew Schwartz. 2022.
\newblock \href {https://doi.org/10.18653/v1/2022.findings-acl.52} {Human language modeling}.
\newblock In \emph{Findings of the Association for Computational Linguistics: ACL 2022}, pages 622--636, Dublin, Ireland. Association for Computational Linguistics.

\bibitem[{Soni et~al.(2024)Soni, Schwartz, Sedoc, and Balasubramanian}]{soni-etal-2024-large}
Nikita Soni, H.~Schwartz, Jo{\~a}o Sedoc, and Niranjan Balasubramanian. 2024.
\newblock \href {https://aclanthology.org/2024.naacl-long.477} {Large human language models: A need and the challenges}.
\newblock In \emph{Proceedings of the 2024 Conference of the North American Chapter of the Association for Computational Linguistics: Human Language Technologies (Volume 1: Long Papers)}, pages 8631--8646, Mexico City, Mexico. Association for Computational Linguistics.

\bibitem[{Volkova et~al.(2013)Volkova, Wilson, and Yarowsky}]{volkova_exploring_2013}
Svitlana Volkova, Theresa Wilson, and David Yarowsky. 2013.
\newblock \href {https://aclanthology.org/D13-1187} {Exploring {Demographic} {Language} {Variations} to {Improve} {Multilingual} {Sentiment} {Analysis} in {Social} {Media}}.
\newblock In \emph{Proceedings of the 2013 {Conference} on {Empirical} {Methods} in {Natural} {Language} {Processing}}, pages 1815--1827, Seattle, Washington, USA. Association for Computational Linguistics.

\bibitem[{Zhong et~al.(2021)Zhong, Tang, Wang, Yin, and Duan}]{zhong_useradapter_2021}
Wanjun Zhong, Duyu Tang, Jiahai Wang, Jian Yin, and Nan Duan. 2021.
\newblock \href {https://doi.org/10.18653/v1/2021.findings-acl.129} {{UserAdapter}: {Few}-{Shot} {User} {Learning} in {Sentiment} {Analysis}}.
\newblock In \emph{Findings of the {Association} for {Computational} {Linguistics}: {ACL}-{IJCNLP} 2021}, pages 1484--1488, Online. Association for Computational Linguistics.

\end{thebibliography}

\appendix
\section{Appendix}
\label{sec:appendix}

\subsection{Pre-training GRIT}
\label{PT_grit_appdx}
\paragraph{Pre-training data.}
We use a subset of the pre-training data for \hart, consisting of the demographics and personality information. This subset contains the Facebook posts from \citet{park_automatic_2015} as used by \citeauthor{soni-etal-2022-human}. Our dataset is consistent with the inclusion criteria for \hart to ensure moderate language history for each user: we include English posts from users with at least 50 total posts and at least 1000 words. This dataset consists of just over 63,000 unique users, which we split into a training dataset consisting of messages from 56,930 users, a development dataset that consists of messages from 1836 users that were not part of the training set, and a test set of messages from a separate set of 4438 users that are neither in training nor the development set. To evaluate the human attribute prediction in \nameope, we use a subset of the test set consisting of messages from 1745 users to accommodate for questionnaire reliability. We use the Facebook posts for the \hulm task and the demographic and personality scores of consenting Facebook users \citep{Kosinski5802} for the human attribute prediction task. 

\begin{table}
    \begin{center}
    \begin{tabularx}{0.49\textwidth} {
    | >{\hsize=0.8\hsize\raggedright\arraybackslash}X
    | >{\hsize=0.8\hsize\centering\arraybackslash}X 
    | >{\hsize=0.8\hsize\centering\arraybackslash}X 
    | >{\hsize=1.3\hsize\centering\arraybackslash}X 
    | >{\hsize=1.3\hsize\centering\arraybackslash}X |
    }
    \hline \bf Age bucket & \#Users & \it{\hart} (Ind) & \it{\nameage} (Ind+Grp) & \it{\nameope} (Ind+Grp) \\ 
    \hline
    
    <18  & 503 & 0.627 & \bf 0.644 & 0.618  \\
    18-21 & 560 & 0.557 & \bf 0.608 & 0.592  \\
    21-30 & 563 & 0.715 & \bf 0.741 & 0.738  \\
    30-45 & 249 & 0.594 & \bf 0.669 & 0.667  \\
    45+ & 68 & 0.567 & 0.546 & \bf 0.599 \\
    
    \hline
    \end{tabularx}
    \end{center}
    \caption{ 
        Disattenuated pearson $r$ for openness over five age buckets using different types of human contexts for error analysis. Bold indicates best in row. 
        }
        \label{tab:error-analy-ope-r}
\end{table}

\begin{table}
    \begin{center}
    \begin{tabularx}{0.49\textwidth} {
    | >{\hsize=0.8\hsize\raggedright\arraybackslash}X
    | >{\hsize=0.8\hsize\centering\arraybackslash}X 
    | >{\hsize=0.8\hsize\centering\arraybackslash}X 
    | >{\hsize=1.3\hsize\centering\arraybackslash}X 
    | >{\hsize=1.3\hsize\centering\arraybackslash}X |
    }
    \hline \bf Age bucket & \#Users & \it{\hart} (Ind) & \it{\nameage} (Ind+Grp) & \it{\nameope} (Ind+Grp) \\ 
    \hline
    
    <18  & 1113 & \text{\doublespace{}\doublespace{}4.07} & \textbf{\doublespace{}\doublespace{}2.52} & \text{\doublespace{}\doublespace{}2.82} \\
    
    18-21 & 1387 & \text{\doublespace{}\doublespace{}6.52} & \text{\doublespace{}\doublespace{}4.00} & \textbf{\doublespace{}\doublespace{}3.89} \\
    
    21-30 & 1557 & \text{\doublespace{}17.82} & \textbf{\doublespace{}12.64} & \text{\doublespace{}13.11} \\
    
    30-45 & \text{\doublespace{}695} & \text{\doublespace{}48.59} & \textbf{\doublespace{}39.79} & \text{\doublespace{}40.43} \\
    45+ & \text{\doublespace{}248} & \bf 114.92 & 121.66 & 134.72 \\
    
    \hline
    \end{tabularx}
    \end{center}
    \caption{
        Mean squared error for age over five age buckets using different types of human contexts for error analysis. Bold indicates best in row (lower error is better). 
        }
        \label{tab:error-analy-age-mse}
\end{table}

\begin{table}
    \begin{center}
    \begin{tabularx}{0.49\textwidth} {
    | >{\hsize=0.8\hsize\raggedright\arraybackslash}X
    | >{\hsize=0.8\hsize\centering\arraybackslash}X 
    | >{\hsize=0.8\hsize\centering\arraybackslash}X 
    | >{\hsize=1.3\hsize\centering\arraybackslash}X 
    | >{\hsize=1.3\hsize\centering\arraybackslash}X |
    }
    \hline \bf Age bucket & \#Users & \it{\hart} (Ind) & \it{\nameage} (Ind+Grp) & \it{\nameope} (Ind+Grp) \\ 
    \hline
    
    <18  & 503 &  0.423 & \bf 0.410 & 0.429  \\
    18-21 & 560 & 0.496 & \bf 0.487 & 0.506  \\
    21-30 & 563 & 0.429 & \bf 0.380 & 0.381  \\
    30-45 & 249 & 0.578 & \bf 0.489 & 0.489  \\
    45+ & \text{\doublespace{}68} & 0.584  &  0.501 & \bf 0.467 \\
    
    \hline
    \end{tabularx}
    \end{center}
    \caption{
        Mean squared error for openness over five age buckets using different types of human contexts for error analysis. Bold indicates best in row (lower error is better). 
        }
        \label{tab:error-analy-ope-mse}
\end{table}

\paragraph{Training.} We use \hart's pre-trained weights as the base weights for \name and randomly initialize the newly introduced weights for human attribute prediction. \name is trained on our pre-training dataset using the 5e-5 learning rate after experimenting with a few learning rates, including that used for \hart's pre-training. Following \hart, and due to computing limitations, each training instance is capped to 8 blocks of 1024 tokens each, with train batch size as 1 per device and evaluation batch size as 20 per device, trained over 2 GPUs for eight epochs. We explored multiple joint losses before resorting to the homoscedastic loss computation. Since \hart caps to 4 train blocks for user-level downstream tasks, we also pre-train \nameage and \nameope with four training blocks.

\subsection{Experimental Settings}
\label{expmt_settings}
We closely follow the experimental settings from \citet{soni-etal-2022-human} and similarly use Optuna framework \citep{akiba2019optuna} for hyperparameter search. We search for learning rates between 5e-6 and 5e-4, and between 1e-7 and 1e-5 for different tasks. We will make our best found hyperparameter values publicly available with our code and results in the github repository. All experiments are run on NVIDIA RTX A6000 GPUs of 48GB. Pre-training takes approx 14 hours for 1 epoch and fine-tuning takes approx 1-4 hours depending on the task.

\end{document}